\documentclass[runningheads]{llncs}
\usepackage[T1]{fontenc}
\usepackage{graphicx}
\usepackage{hyperref}
\usepackage{mathtools}
\usepackage{enumitem}
\usepackage{subcaption}

\begin{document}
\title{Enhancing FKG.in: automating Indian food composition analysis\thanks{Supported by Ashoka Mphasis Lab}}
%
%
\author{Saransh Kumar Gupta\inst{1}\orcidID{0009-0000-5887-2301} \and
Lipika Dey\inst{1}\orcidID{0000-0003-3831-5545} \and
Partha Pratim Das\inst{1}\orcidID{0000-0003-1435-6051} \and
Geeta Trilok-Kumar\inst{1}\orcidID{0000-0001-8442-2881} \and
Ramesh Jain\inst{2}\orcidID{0000-0003-2373-4966}}
\authorrunning{S. K. Gupta et al.}
%

\institute{Ashoka University, India \\
\email{\{saransh.gupta, lipika.dey, partha.das, geeta.kumar\}@ashoka.edu.in}
\url{https://www.ashoka.edu.in/}
\and
Institute of Future Health, UC Irvine, USA \\
\email{jain@ics.uci.edu}\\
\url{https://futurehealth.uci.edu/}}
\maketitle              

\begin{abstract}
This paper presents a novel approach to compute food composition data for Indian recipes using a knowledge graph for Indian food (FKG.in) and LLMs. The primary focus is to provide a broad overview of an automated food composition analysis workflow and describe its core functionalities: nutrition data aggregation, food composition analysis, and LLM-augmented information resolution. This workflow aims to complement FKG.in and iteratively supplement food composition data from verified knowledge bases. Additionally, this paper highlights the challenges of representing Indian food and accessing food composition data digitally. It also reviews three key sources of food composition data: the Indian Food Composition Tables, the Indian Nutrient Databank, and the Nutritionix API. Furthermore, it briefly outlines how users can interact with the workflow to obtain diet-based health recommendations and detailed food composition information for numerous recipes. We then explore the complex challenges of analyzing Indian recipe information across dimensions such as structure, multilingualism, and uncertainty as well as present our ongoing work on LLM-based solutions to address these issues. The methods proposed in this workshop paper for AI-driven knowledge curation and information resolution are application-agnostic, generalizable, and replicable for any domain.

\end{abstract}
\section{Introduction}

Food is a fundamental necessity of life and greatly determines its quality. A healthy relationship with food develops when people enjoy their favorite dishes without compromising their health. However, too often, there is a wide gap between what people love to eat and what they are advised to consume for their well-being. This discrepancy can be traced to humans' multidimensional relationship with food, shaped by their evolutionary, social, and economic history. It reflects a view of food that extends beyond its role as a mere energy source and essential nutrients for sustenance. Historically, food choices have been largely affected by availability (climate, geography, water proximity), accessibility (political structures, legal frameworks, budget constraints), trade (industrialization, cultural exchange, economic status), customs (social hierarchies, religious practices, festivals), and technology (processing, preservation, transportation). However, with rising lifestyle diseases, there is a need to refocus on eating habits that support both mental and physical wellness. For holistic well-being, understanding the nutritional composition of preferred foods is essential. This has spurred the development of health applications with extensive food knowledge bases tailored to diverse ethnic groups.

While this tragedy of food affects most of the world, the situation in India is especially dire, with diverse malnutrition issues ranging from stunting and undernourishment to overweight and obesity, compounded by widespread nutritional insecurity \cite{nguyen2021}. Eating right remains a largely neglected challenge for most of the Indian population i.e. approximately one-fifth of the world’s total. This is partly due to the country's historical struggle with severe food insecurity, where the focus has largely been on eating sufficiently rather than eating right. However, the larger issue lies in the absence of a consolidated, dynamic, and accessible knowledge base encompassing various aspects of Indian food. Although traditional Indian food, especially Indian meals, has a reputation for being balanced and sustainable, much of this knowledge remains undocumented or inadequately recorded. Additionally, there is a lack of comprehensive food composition tables necessary for accurate food composition analysis. As a result, most health applications used by the Indian population are unable to provide reliable, personalized, and balanced recommendations that cater to the Indian palette.

In this paper, we present our work on building a food knowledge graph using Indian recipes curated from diverse sources, enhancing it with nutrient information, and building an automated food composition analysis workflow. Engineering such a knowledge graph and sourcing information from recipe websites and cookbooks involves addressing several linguistic challenges - including multilingualism - such as entity and relationship name resolution, structural differences in the recipe information presented across sources, and uncertainties in it. Other major challenges arise from the gaps and incompleteness of food composition tables owing to the vast diversity and complexity of Indian food, inadequate recipe documentation of recipes, and the high cost of analytical experiments to empirically generate food composition data. This paper builds upon our earlier work \cite{Gupta24}, which introduced the knowledge graph (FKG.in) designed to store Indian recipes and their ingredients alongside numerous descriptive properties.

The novelty of the proposed work lies in the innovative use of Large Language Models (LLMs) to address structural and multilingual challenges, as well as the inherent uncertainties in recipes. A series of prompts are directed at the LLM, with the recipe provided as context, to extract reliable information about recipes, ingredients, their descriptors, and their measurements. This information is combined with food composition tables to determine the nutritional composition of recipes. While we have observed that the quality of results obtained can be further improved by incorporating additional details such as cooking techniques, cookware, and cooking time, the information curated thus far already provides a valuable source for knowledge discovery about Indian recipes. Additionally, we demonstrate how this knowledge graph can be leveraged to generate diet-based health recommendations and recipe suggestions.

The rest of the paper is organized as follows. Section 2 presents an overview of the work done in food knowledge graphs and personal digital health applications. Section 3 presents the unique challenges around consolidating and estimating nutritional information for Indian food, especially recipes. Section 4 briefly describes FKG.in for additional context. Section 5 compares three key sources of food composition data for Indian cuisine, their strengths, and gaps. Section 6 describes the Nutrition Data Aggregator (NDA) agent. Section 7 presents the challenges in analyzing recipe information along the lines of structure, multilingualism, and uncertainty and details the LLM-Augmented Information Resolution (LAIR) agent and its attempts to solve some of these challenges. Section 8 describes the automated food composition workflow and the Food Composition Analysis (FCA) agent. Section 9 presents some results. Section 10 concludes with a note on a few limitations of this work and future directions.

\section{Related Work}

Recently, individuals in urban regions in India (as in many other countries) have become more conscious of the nutritional composition of their food and have actively started adopting healthier lifestyles. This shift is driven by growing awareness of physical and mental well-being, as well as the rising prevalence and escalating threat of diet-related diseases such as obesity, hypertension, and diabetes in India \cite{vennu2019prevalence}. This is increasingly being facilitated through technology and the application of scientific knowledge about food, health, nutrition, and behavior wherein notions like Personalized Health Navigators (PHN) \cite{cooper2024phn,iso_9472_10000,nag2019navigational} and Personalized Digital Health (PDH) \cite{iso_11147_2023,jain2023b} have emerged as key areas for research, development, and practical implementation.


Food composition refers to the nutritional information of food items, including macronutrients (carbohydrates, proteins, fats), micronutrients (vitamins, minerals), and other essential food components (fiber, water, bioactive compounds). Several applications and APIs, such as Edamam \cite{edamam}, Fitterfly \cite{fitterfly}, Healthify Me \cite{healthifyme}, My Fitness Pal \cite{myfitnesspal}, Nutritionix\cite{nutritionix}, Poshan Atlas \cite{poshanatlas}, and USDA FoodData Central \cite{usda_fooddata_central} already provide nutritional information to their users. However, advances in food composition analysis technologies, personalized dietary recommendations, and the adoption of PHN/PDH are possible only with a digital representation of food, nutrients, and all their constituents. Hence, research at the intersection of food and health is gaining importance in areas like dietary management, precision nutrition, food safety, and food anthropology.

Many food ontologies and food knowledge graphs have been developed to support a variety of food computing applications \cite{Min19,Min20,Min22}. However, with the ubiquitous rise of unreliable and unverified information on the Internet - further amplified by chatbots built on LLMs - it has become imperative to combat food and nutrition misinformation using the latest technological advances. In response, several initiatives have emerged to address issues related to food, nutrition, and health such as PIPS \cite{Cantais05}, FoodOn \cite{Dooley18}, FoodKG \cite{Haussmann19}, FOODS \cite{Snae08}, and WikiFCD \cite{Thornton2021IntroducingWM}. In the Indian context, notable contributions include the Indian Food Composition Tables \cite{Longvah17}, the Indian Nutrient Databank \cite{Vijayakumar24}, \textit{Nutritional Profile Estimation in Cooking Recipes} \cite{Kalra2020}, and \textit{Dish detection in food platters} \cite{goel2023dishdetectionfoodplatters}. Building on this momentum, FKG.in \cite{Gupta24} represents our effort to build a comprehensive, reliable, and granular knowledge graph for Indian food with associated intelligence, automation, and scalability to address some of these challenges.

\section{Challenges in Indian Food Composition Analysis}

India is a country of remarkable diversity. Climate, topography, culture, language, agricultural practices, availability of local ingredients, and economic conditions vary significantly every few kilometers, distinctly shaping food consumption patterns. It is also home to a vast array of cuisines, where the defining characteristics are often rooted in geographical, historical, religious, or economic contexts. Recipes from various Indian cuisines reflect some of these characteristics but not sufficiently. Designing an ontology and building a knowledge graph for Indian food necessitates careful consideration of these and other factors. Below, we outline key challenges in Indian food composition analysis:

\begin{enumerate}
    \item \textbf{Structure:} Indian recipes typically lack a standard format for describing ingredients, their measures, size, state of processing, form, etc. This inconsistency makes it difficult to extract essential details and descriptors of ingredients accurately, which are crucial for precise food composition analysis.
    \item \textbf{Multilingualism:} Ingredients are known by different names across the country, with recipes using English, vernacular, and colloquial terms. These names may appear in Roman or Indian scripts and often have varying phonetic spellings, regional dialects, and code-mixing. Such variations complicate straightforward lookups in static food composition tables. Additionally, homonyms and semantic ambiguities, such as \textit{saag} meaning different greens across regions, require context-aware resolution in food composition analysis.
    \item \textbf{Uncertainty in Ingredient Measures:} Indian cooking often eschews precise measurements, relying instead on ambiguous terms (e.g., one small \textit{katori} (bowl), 2 large spoons). Sometimes, no measure is specified for common ingredients and spices assumed to be familiar and regularly used (e.g., ginger-garlic paste, \textit{jeera} (cumin) powder). Such imprecision greatly impacts nutrient computation when key ingredient units and quantities are missing.
    \item \textbf{Uncertainty in Nature of Ingredients:} Since recipes often lack standardized formats or guidelines, they frequently list generic ingredients (e.g., 1 cup oil for frying) rather than specifying their particular type. As different oils have different compositions, this also affects nutrition calculations. In this work, default labels are applied wherever possible, based on common knowledge of Indian cooking and cuisine. E.g., the knowledge that mustard oil is used for cooking fish in \textit{Bengali} cuisine or groundnut oil is used in \textit{Gujarati} cuisine can be assumed in the absence of more specific details.
\end{enumerate}

\section{FKG.in: A Knowledge Graph for Indian Food}


Indian food, especially meals, primarily consists of cooked dishes. Hence, recipes serve as its core and are instrumental in building the Indian food knowledge graph (FKG.in) \cite{Gupta24}. FKG.in draws inspiration from FoodOn \cite{Dooley18} and FoodKG \cite{Haussmann19}, adapting these frameworks wherever necessary to suit the Indian context. It aims to capture essential properties of Indian food in terms of culinary language, cooking variations, and precision nutrition, including, but not limited to, elements such as meals, recipes, ingredient details, cooking methods, cookware, and dietary labels. The design principles followed a modular and flexible approach to knowledge curation. To address linguistic challenges including multilingualism as well as inherent uncertainties in Indian recipes across diverse sources, we employed LLMs quite extensively. In the following sections, we discuss three verified food composition data sources and how FKG.in is enhanced by incorporating nutritional information from them through an automated workflow.

\section{Sources of Nutritional information for Indian Food}

Since we could not find a single comprehensive and open-access source for obtaining nutritional information for Indian food items, we have used multiple sources as listed below along with their unique characteristics, strengths, and gaps:

\begin{enumerate}
    \item \textbf{Indian Food Composition Table (IFCT, 2017):} This comprehensive resource \cite{Longvah17} was developed by the \textit{Indian Council of Medical Research-National Institute of Nutrition} (ICMR-NIN) in 2017 and provides detailed information about the nutritional composition of various food items commonly consumed in India. It also provides scientific names, food groups, dietary tags (e.g., vegetarian, eggetarian), over 150 nutrient and food component data points for 528 food items, and common names in 18 Indian vernacular languages for many of them. However, while it includes a wide array of food items, it remains inconsistent and incomplete. E.g., 3 potato variants are found in IFCT viz. ‘potato, brown skin, big’, ‘potato, brown skin, small’, and ‘potato, red skin’ but the food composition data for ‘boiled potatoes’ is missing. Similarly, both regular and roasted variants of vermicelli are found in IFCT but the roasted variant for groundnuts and the milled variant of Bengal gram i.e. \textit{besan} (Bengal gram flour) is missing. IFCT also misses out on some crucial ingredients that are commonly used in modern India such as cheese, tofu, butter, mayonnaise, noodles, broccoli, salt, etc., and misses some common nutrient data points such as Iodine and Vitamin B12. For all the oils and \textit{ghee} (clarified butter), IFCT does not provide information about  Energy, Cholesterol, and many important micronutrients.
    
    \item \textbf{Indian Nutrient Databank (INDB, 2024):} This data repository \cite{Vijayakumar24} was built by \textit{Anvaad Solutions} in 2024 and it builds on IFCT-2017 to fill in many of its gaps. For instance, it includes broccoli, roasted groundnuts, \textit{besan}, cheese, noodles, salt, etc. It brings in food composition data for both raw ingredients and their variants from multiple sources viz. ICMR-NIN IFCT-2004 and nutrient databases from the United Kingdom \cite{uk2015composition} and the United States \cite{usda_fooddata_central} in that order of priority making the total number of individual food items available in INDB 1095. INDB has also curated a vocabulary of common measurement units for both ingredients (e.g., tablespoon, pinch, sprig) and recipes (e.g., plate, bowl, slice) and their mapping to weights in grams for some of these common units. Notably, INDB also includes many complex ingredients such as \textit{garam masala} (a blend of ground spices), tomato ketchup, and \textit{rasam} (tamarind-tomato soup) powder which themselves follow recipe instructions and provides a detailed food composition table for 1014 unique and ‘standard' Indian recipes which were sourced from 2 books on Indian cooking and selected recipe blogs. However, an issue with the INDB database is that many nutrient data points are not available for all the items. Only 38 out of the IFCT-2017’s 150+ data points feature in INDB for all 1095 ingredients. INDB also does not include food composition information about branded or packaged food items yet. Additionally, while INDB provides some local names for recipes, it does not provide the scientific names, food groups, dietary tags, and vernacular names for the newly added ingredients.
    
    \item \textbf{Nutritionix (2010):} This nutrition database \cite{nutritionix} offers food composition details for thousands of branded, restaurant, and generic food items. While it caters primarily to the food items in the U.S., it includes a reasonable number of Indian food items as well, likely owing to the popularity of Indian cuisine. Most notably, it includes processed, form, and size variants of several ingredients that IFCT and INDB do not sufficiently include. E.g., chopped spinach and minced garlic are found in addition to spinach and garlic as well as boiled potato and steamed tofu are found in addition to potato and tofu in the Nutritionix database. It is important to note here that Nutritionix may not always give the most accurate results for Indian food as it is primarily built on top of USDA whereas the nutritional content of food varies depending on the geographic and climate conditions in which it is produced. In this light, only the IFCT data captures the nutrient information of Indian food items accurately as it compositely samples each food item from six different regions covering the entire country and averages them \cite{Longvah17}.
\end{enumerate}

    The above discussion highlights the uniqueness and inadequacies of IFCT, INDB, and Nutritionix. In the next section, we present an automated workflow designed to dynamically aggregate the nutrient information for Indian food items (ingredients and recipes) from them. The proposed workflow enhances the previously described knowledge graph, FKG.in, by incorporating the food composition data into it at scale. Additionally, it creates an enhanced food composition table by consolidating the information from three reliable sources.

\section{Nutrition Data Aggregation}

\vspace{-0.09em}

In this section, we describe the design of the Nutrition Data Aggregator (NDA) agent, that enhances FKG.in with an aggregated food composition table, named  FKG.in-FCT. For any recipe, once its ingredient names and measurements are extracted from the source, the NDA agent checks IFCT, INDB, and Nutritionix, in that order, to obtain nutritional information from them, if available. All ingredients newly encountered are added to FKG.in-FCT, resulting in a consolidated food composition table, that is also enriched with multilingual information about the ingredients. For each ingredient, it also stores information about its different variants along the axes of forms (e.g., seeds, powder, or paste), processing steps (e.g., boiled, roasted, or steamed), and size (e.g., small, medium, large). When an ingredient is missing from both IFCT and INDB, an API call to Nutritionix adds both the ingredient as well as its nutrient information to FKG.in-FCT. This allows FKG.in-FCT to include various ingredient variants commonly used in Indian recipes but not present in IFCT and INDB, significantly expanding the list of ingredients. The enhanced FKG.in-FCT ensures uniformity of scale across sources with the appropriate unit and serving size conversions.

Admittedly, significant limitations remain in achieving precise nutritional estimates. Firstly, the number of ingredients (with variants) used in Indian cuisine is countable but the number of recipes is virtually infinite, making a detailed analysis of numerous cooked samples impractical, especially in India’s diverse culinary context. Secondly, many of Nutritionix’s data points are accurate for American variants of ingredients but can only approximate their Indian counterparts. Additionally, in rare cases, approximations may arise because of the form in which an ingredient is used in a recipe. E.g., singhara (water chestnut) flour is missing from all three sources, so the nutrient values of the fruit are used as a proxy. Lastly, cooking is known to reduce nutrients, yet due to the lack of data on retention and yield factors for Indian ingredients and cooking methods, such adjustments could not be performed. The proposed workflow streamlines food composition analysis for the rapidly growing and continuously evolving repertoire of Indian recipes in FKG.in, adopting a trade-off in precision to enable large-scale automated calculations. Some of the challenges mentioned above, along with managing missing food items and leveraging LLMs to resolve uncertainties, are explored further in the next section.

\section{LLM-Augmented Information Resolution}

Despite using three reliable food composition tables, several ingredients used in Indian recipes could not be found in them. In some cases, ingredients were present in the databases but could not be found easily due to linguistic variations. Additionally, certain Indian food names could not be directly mapped to their equivalents in Nutritionix. LLMs were employed as a tool to navigate these difficulties. While it is well-known that LLMs may hallucinate, and our experience confirms that they cannot always be fully trusted for nutrient information, we assert that LLMs can play a crucial role in mitigating challenges of extracting information from unstructured recipe texts, resolving food or ingredient name ambiguities arising from multilingualism, and addressing quantitative uncertainties when recipe units and ingredient measures lack specificity and precision. With appropriate prompts, LLMs proved to be surprisingly effective at resolving many of these issues. We have developed an LLM-Augmented Information Resolution (LAIR) agent using OpenAI’s GPT-3.5 Turbo\footnote{https://platform.openai.com/docs/models} to process the recipe and ingredient information, leveraging prompt engineering to fine-tune the results as needed. The LAIR agent employs LLMs for both pre-processing as well as post-processing tasks to effectively carry out the following activities:   

\begin{enumerate}
    \item \textbf{Recipe Language Translation:} We collect recipes from a range of sources including recipe blogs, cookbooks, and Wikipedia, and also accept them as user inputs. Some of these recipes are written in vernacular Indian scripts such as \textit{Hindi}, \textit{Bengali}, or \textit{Tamil} and require transliteration to Roman script for their assimilation into the FKG.in. The LAIR agent uses LLMs for transliteration as well as translation and gets reasonably good results. 
    
    \item \textbf{Recipe Format Normalization:} Recipe sources use varying formats for organizing recipes wherein cooking instructions may be presented and formatted differently, attributes may be mentioned using inconsistent or missing labels, and the recipe descriptions may contain information that could be useful for food composition analysis. We use LLMs to extract relevant information from recipes such as ingredient details, cooking instructions, category tags, cooking time, etc., and normalize them as per a standard format in JSON for them to be added to FKG.in in a streamlined manner.
    
    \item \textbf{Ingredient Format Normalization:} Recipe sources also vary in detailing the list of ingredients. E.g., ‘2 cups boiled \textit{aloo} (potatoes) (medium-sized), chopped' and ‘½ kg chopped medium potatoes to be taken after boiling them' refer to the same ingredient referent approximately\footnote{Assuming a cup is 240 ml and ignoring the ingredient density variations.}. In this step, LLMs are instructed to singularize, normalize and segment the ingredient details such as their names, variant descriptors, quantities, and units, and store them in a unified JSON format as \{"ingredient": "potato", "form": "chopped", "process": "boiled", "size": "medium", "quantity": "2", "unit": "cups"\}.

    \item \textbf{Ingredient Name Resolution:} Multilingualism leads to the same ingredients being mentioned across various sources with different identifiers. E.g., a potato may be mentioned as \textit{alu} (\textit{Hindi}), \textit{bateka} (\textit{Gujarati}), and \textit{oalu} (\textit{Kashmiri}). Quantitative variations like ‘whole ginger’, and ‘an inch of ginger' are also encountered, which add to the difficulty of food composition analysis. LLM helps us effectively remove the redundancies and normalize the non-vernacular aliases to the preferred label while ensuring that the important descriptors of form, processing, and size are not removed in the process.
    
    \item \textbf{Ingredient Category Assignment:} FKG.in stores ingredients in hierarchical subcategories such as RootOrTuberousVegetable, MeatFromChicken, ProcessedMushroom, etc. For every unique ingredient, we infer their potential categories as per our ontology design using LLMs. E.g., potato will be assigned the category RootOrTuberousVegetable (Ingredient -> PlantOriginFood -> PrimaryFoodCommodityOfPlantOrigin -> Vegetable -> RootOrTuberousVegetable) whereas mushroom pickle will be assigned the category ProcessedMushroom (Ingredient -> FungusOrigin -> SecondaryFoodCommodityOfFungusOrigin -> ProcessedMushroom).
    %
    
    \item \textbf{Ingredient Unit Normalization:} Often same units are referred to by different identifiers. E.g., a tablespoon is commonly found to be written across sources such as \textit{tablespoons}, \textit{TABLESPOON}, \textit{T.}, \textit{TB.} \textit{tbsp.}, \textit{Tblsp.}, \textit{tbs.}, \textit{tbl.}, \textit{tbls.}, \textit{a large spoon}, etc. We prompt the LLM to normalize ingredient units by converting them to lowercase, singular form, and standard full names, exemplifying how all the above units should be normalized to \textit{tablespoon}.

    \begin{figure}[htbp]
        \begin{minipage}[b]{0.48\textwidth}
            \centering
            \fbox{\includegraphics[width=\textwidth]{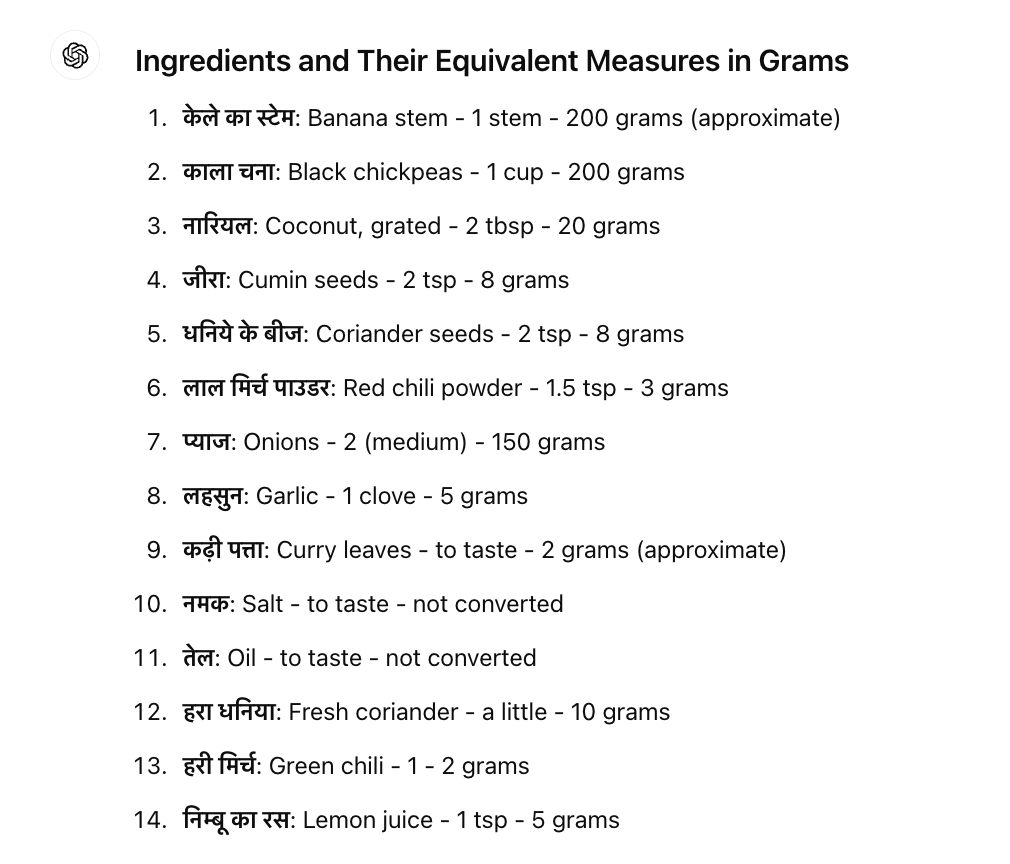}}
            \caption{A sample ChatGPT response to a user prompt for translating ingredient names from Hindi to English and estimating their weights in grams}
            \label{fig:chatgpt_response_for_translation_and_weight_estimation}
        \end{minipage}
        \hfill
        \begin{minipage}[b]{0.48\textwidth}
            \centering
            \fbox{\includegraphics[width=\textwidth]{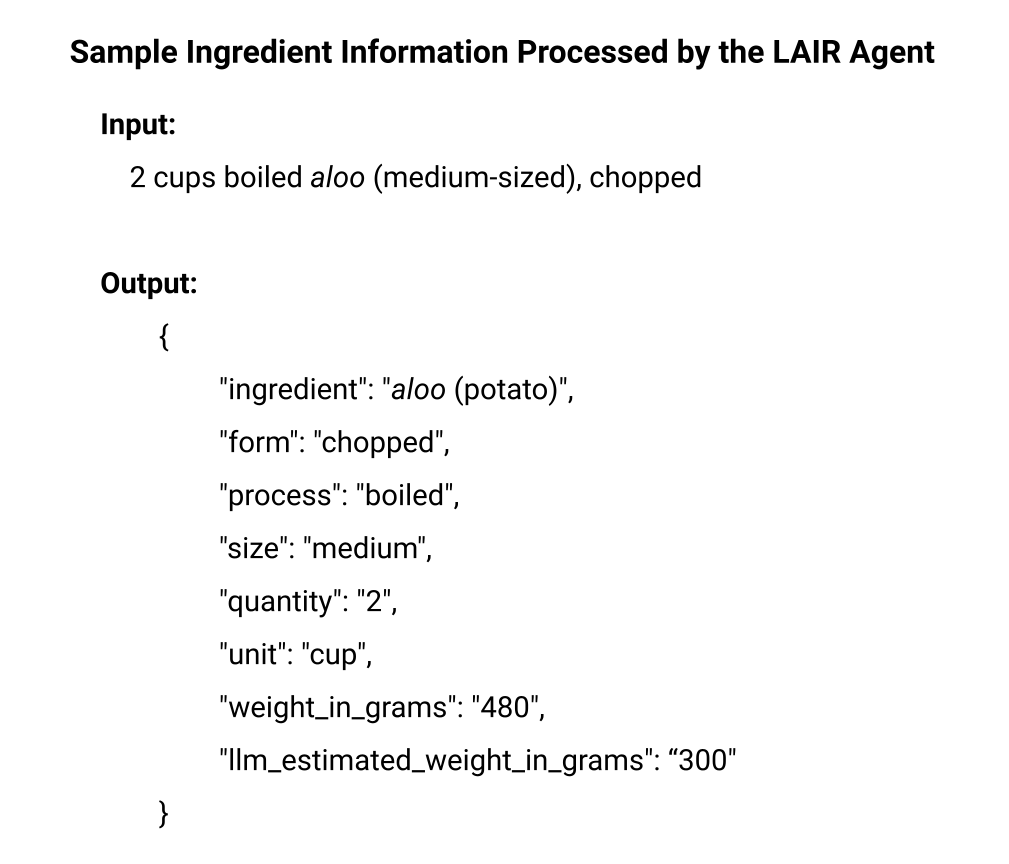}}
            \caption{A sample LAIR agent output for segmenting, formatting, and normalizing ingredient information, and estimating its weight in grams}
            \label{fig:lair_agent_output_for_processing_ingredient_information}
        \end{minipage}
    \end{figure}

    \item \textbf{Ingredient Measurement Resolution:} Often the units mentioned are associated with specific ingredients and refer to a specific weight per unit. E.g., garlic is commonly mentioned with the units ‘clove’ or ‘segment’ (3-7 grams), as well as ‘bulb’ or ‘head’ (40-80 grams) depending on the garlic size. On the other hand, certain units apply to multiple ingredients. E.g., ‘pods’ are frequently used with cardamom, vanilla, and tamarind. To resolve such issues with ingredient units, we use a combination of \textit{Dietary Measurement Ontology} that contains mapping rules built on top of INDB’s unit vocabulary and complemented by LLM. This helps us store standard unit conversions (e.g., 1 teaspoon = 5g, 1 cup = 48 teaspoon), map ingredient and variant-specific units (e.g., bulb/head for a whole garlic, clove/segment for a single section of garlic), and translate ingredient variant-specific units to other units (e.g., 1 bulb of garlic = 10 cloves, 1 clove of garlic (minced) = 1 teaspoon,). Such mapping rules help us convert both standard and inexact units to their respective weights in grams dynamically.  
    We also use the unit information alongside ingredients in the recipes to extend the unit vocabulary by iteratively validating it with a human-in-the-loop. Occasionally, when an ingredient quantity is mentioned as a range such as 2-4 cups or 4-6 cloves, we average it, and if required, we convert the fractions to their decimal counterparts for smoother processing. It is important to note that when the recipe mentions the exact ingredient weight alongside measurements, we use it directly as it is already resolved. The running example JSON gets updated to \{"ingredient": "potato", "form": "chopped", "process": "boiled", "size": "medium", "quantity": "2", "unit": "cup", "weight\_in\_grams": "480"\}.
    
    \item \textbf{Ingredient Weight Estimation:} Density is commonly ignored while calculating the weight of an ingredient in grams wherein the measurement containers often use milliliters and assume a convenient translation of 1gm = 1mL for all ingredients, including solids. We have found a more accurate representation in \cite{archanaskitchenweightsmeasurement} but only for a few ingredients. The lack of a more complete data source often makes it difficult to estimate the weight of an ingredient from both the standard (cup, tablespoon) as well as inexact (glass, handful) measurement units to grams. We also use LLMs to estimate the weight in grams for each ingredient based on their quantity and unit such that the example JSON mentioned above gets updated to \{"ingredient": "potato", "form": "chopped", "process": "boiled", "size": "medium", "quantity": "2", "unit": "cup", "weight\_in\_grams": "480", "llm\_estimated\_weight\_in\_grams": "300"\}. We have observed that LLM mostly estimates this weight reasonably well by taking both the density and retention factor of the ingredient implicitly into account although with occasional errors. This helps calculate recipe nutritional information by deriving unit mapping-based weights when possible and using LLM-estimated weights as alternatives when necessary. 
    
    \item \textbf{Recipe Dietary Label Tagging:} In generating and improving diet-based health recommendations, it is often helpful to automatically infer whether certain dietary tags apply to recipes along the axes of dietary practices (e.g., vegetarian, pescatarian), health labels (e.g., keto-friendly, low-sugar), and allergen labels (e.g., contains-dairy, contains-peanuts). We have curated a list of 60+ such tags, formalized them into several rules, and designed a pipeline to assign these rule-based tags to recipes based on the presence, categories, and measurements of ingredients. This step extensively uses the earlier mentioned LLM-assigned category and the calculated or estimated ingredient weights to infer the appropriate dietary tags applicable to a recipe.
    
    \item \textbf{Recipe Latent Information Inference:} Occasionally, it is useful to know which recipes are related and which ingredients are related in addition to knowing which ingredients can substitute other ones to recommend health-based food alternatives. We use LLMs to extract these and other latent details from recipe sources as they are often present in the recipe description and notes but not explicitly mentioned in the recipe cards. We prompt LLM to ensure that the information for these attributes is only extracted from the given recipe source and not inferred from elsewhere to ensure soundness although LLM tends to answer such questions quite well on its own also.
    
\end{enumerate}

\section{Automated Food Composition Analysis Workflow}

\vspace{-2em}
    
\begin{figure}
    \centering
  \includegraphics[scale=0.5]{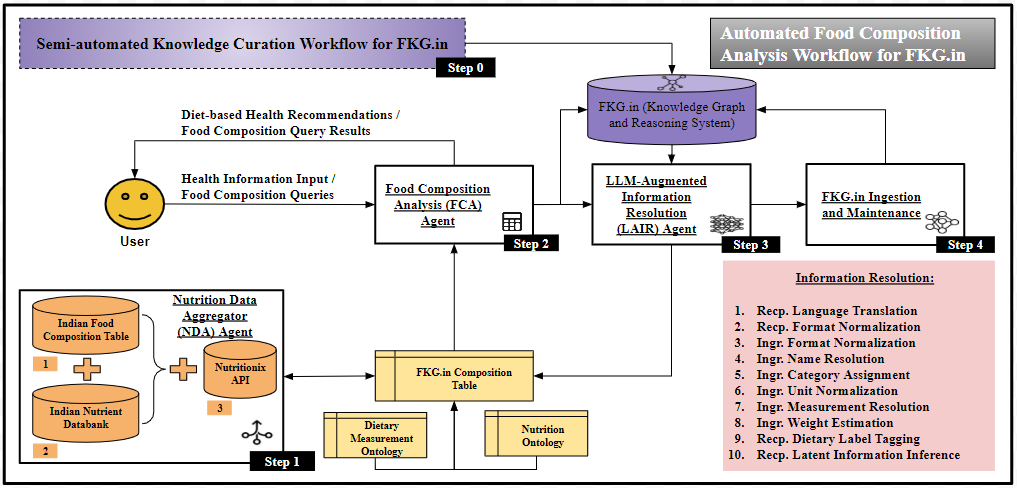}
  \caption{Automated Food Composition Analysis Workflow for FKG.in} \label{fig:food_composition_analysis_workflow}
\end{figure}

Figure \ref{fig:food_composition_analysis_workflow} presents the automated food composition analysis system. At its heart is the Food Composition Analysis (FCA) agent which links the knowledge graph of Indian food i.e. FKG.in, the aggregated Indian food composition table i.e. FKG.in-FCT, and the LLM-Augmented Information Resolution (LAIR) agent, as well as provides an easy-to-use interface for interaction with users. The individual components and their roles in the workflow are explained below:
    \begin{itemize}
        \item \textbf{Step 0: Knowledge Curation -} This includes a set of steps for initializing FKG.in with data from reliable food ontologies, vocabularies, and various sources of recipes. The LLM-augmented information extraction process along with data validation, inconsistency resolution, and data ingestion steps are presented in detail in \cite{Gupta24}. The resulting knowledge graph is stored in an OWL format and implemented using Ontotext's GraphDB. At the end of this workflow, FKG.in stores the list of ingredients with their respective descriptors and measurements for all the recipes exactly as mentioned in their sources. At this point, the recipes do not contain any information about their food composition unless it is explicitly given in the source.
        
        \item \textbf{Step 1: Nutrition Data Aggregation (NDA) Agent -} The NDA agent and the FKG.in-FCT creation process were described in the earlier section. 
        
        \item \textbf{Step 2: Food Composition Analysis (FCA) Agent -} The primary role of the FCA agent is to derive the nutrient and food composition of a recipe based on its ingredients and their quantities. For each dish stored in FKG.in, it uses the FKG.in-FCT to look up ingredients, scale the nutrient values proportionally, and sum them up to calculate the recipe's food composition data. Cooking process transformations are not accounted for at this stage as \cite{Vijayakumar24} found the differences for Indian food with and without retention factors to be small, with notable losses in vitamin C, potassium, and phosphorus.

        The FCA agent is also designed to provide on-demand nutritional information for food items not stored in the knowledge graph. Users can directly interact with the FCA agent via an intuitive interface. For a search involving a dish or recipe without ingredient details, the FCA agent first queries its local knowledge base FKG.in and FKG.in-FCT, using fuzzy matching techniques to find the closest match. It then outputs the nutritional information based on the stored recipe details. A recipe input with details of ingredients and their measures follows the same steps described earlier.

        In addition to dynamically calculating food composition values, the FCA agent provides users with simple, diet-based health recommendations as per \cite{nindietaryguidelines}. These recommendations are tailored based on inputs such as personal information (age, gender, weight, height), physiological stage (infant, child, adolescent, adult, elderly, pregnancy, lactation), activity (type, duration, frequency, intensity, calories burned), dietary preferences (food choices, 24-hour dietary recall, hydration, allergies), and weight goals (gain, lose, maintain).
        
        
        
        \item \textbf{Step 3: LLM-Augmented Information Resolution (LAIR) Agent -} Since recipes are curated from multiple sources, they are quite unstructured. From ingredient names to their units of measurement, nothing is assumed to be available in a standardized way. The LAIR agent works to take in an input, normalize it, handle multilingualism, resolve uncertainties, and obtain weight measurements to calculate nutrient and food composition information. In rare cases, when the FCA agent encounters recipes with ingredients missing from FKG.in-FCT, the LAIR agent also retrieves their nutritional information from public sources, marking it as LLM-sourced in FKG.in-FCT. 
        
        \item \textbf{Step 4: FKG.in Ingestion and Maintenance -} This step ensures that all information obtained for the recipes and ingredients in Step 3, including the ones from public sources, is appropriately ingested in and appended to FKG.in. This is done by following the same soundness assessment and inconsistency resolution approaches that apply to the knowledge curation workflow in Step 0. Since the verification process is manual and time-consuming, there is a time lag between ingesting the information in the knowledge base and its availability for food composition analysis. 
    \end{itemize}

\section{Current Status of FKG.in}

\begin{figure}[]
    \begin{subfigure}[b]{\textwidth}
        \fbox{\includegraphics[width=\textwidth]{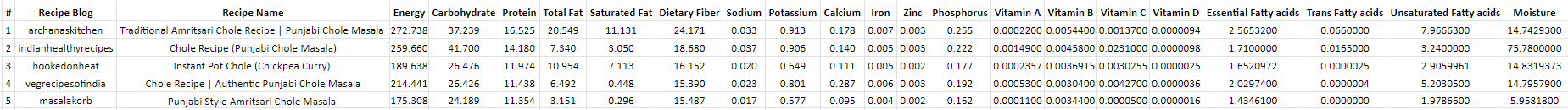}}
        \caption{\textit{chhole masala} (chickpea curry) variants in decreasing order of protein}
    \end{subfigure}
    
    \begin{subfigure}[b]{\textwidth}
        \fbox{\includegraphics[width=\textwidth]{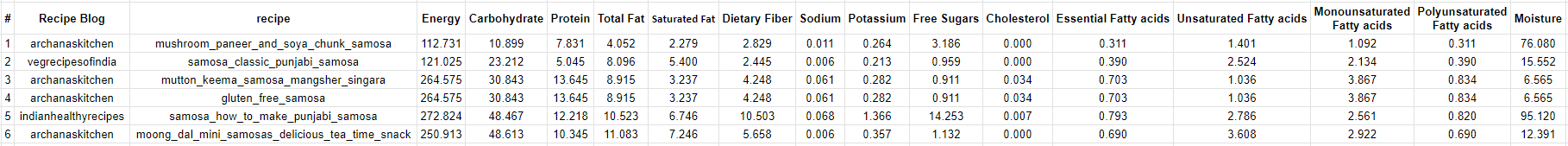}}
        \caption{\textit{samosa} (fried pastry with savory filling) variants in increasing order of total fat}
    \end{subfigure}

    \caption{Energy (kCal) and selected nutrients (g) comparison of recipe alternatives}\label{fig:comparison_of_recipe_alternatives}
\end{figure}

Figure \ref{fig:comparison_of_recipe_alternatives} illustrates a small sample of \textit{chhole masala} (chickpea curry) variants and \textit{samosa} (fried pastry with savory filling) alternatives, stored in FKG.in. These include selected nutrient information, enabling comparisons to support recipe recommendations tailored to dietary preferences, nutritional needs, and health objectives. As can be appreciated here, describing a ‘standard' \textit{samosa} or \textit{chhole masala} is quite difficult as their recipes vary widely across regions, cultures, and preferences. Thus, it highlights the ambiguity in both defining 'standard' recipes for cooked food items as well as the complications in determining the food composition of purported 'standard' recipes. This led us to treat each recipe as a unique instance of the corresponding cooked food item. FKG.in currently includes information on 25,000+ unique recipe instances, sourced from 15+ recipe sites and 5+ cookbooks, with more recipes being added iteratively.

\vspace{-1em}

\vspace{-3em}

\section {Conclusions and Future Work}

In this paper, we have presented the design of an automated workflow for enhancing the Indian food knowledge graph (FKG.in), which already contains an extensive collection of recipes and ingredients, with food composition information. The automated workflow facilitates the expansion of the FKG.in Food Composition Table (FCT) with detailed nutritional data for a vast number of ingredients. Additionally, this process results in an aggregated, validated, and comprehensive dataset of recipe nutrient compositions alongside the addition of new recipes from diverse sources. Most importantly, we demonstrated how the LLM-Augmented Information Resolution (LAIR) agent effectively addresses various structural, linguistic, and uncertainty-related challenges in parsing and analyzing recipe and ingredient information. This proposed approach is application-agnostic, supports dynamic computation of nutritional information for countless recipes, and enhances both the scope as well as the granularity of dietary recommendations.

\vspace{\baselineskip} 

Some limitations of this work and potential future improvements include:

\begin{enumerate}
    \item Verified FCTs from neighboring countries like Bangladesh, Nepal, Pakistan, and Sri Lanka can be incorporated to represent Indian subcontinental cuisine better, scale FKG.in up, and improve FKG.in-FCT's completeness.
    \item Nutritionix is not always accurate for Indian foods, as it relies on USDA data. However, the NDA agent calls the Nutritionix API only if an ingredient is missing from both IFCT and INDB, which occurs infrequently.
    \item Without accounting for cooking retention and yield factors, our nutrition calculations may be slightly overestimated. We have retention factors from the USDA database in FKG.in-FCT but are currently not using them. Future updates to FKG.in will allow us to provide more accurate food composition results as it will utilize in-depth knowledge of Indian cooking styles as well.
    \item The FCA agent doesn’t suggest specific food items based on food group inadequacy. Incorporating dietary tag labels in the recommendation workflow can allow it to recommend food items to address dietary gaps as per \cite{nindietaryguidelines}.
    \item The validation process for the LLM-generated information such as weight estimation, vernacular translation/transliteration, and category assignment needs to be improved. Currently, it is human-dependent and slow.   
\end{enumerate}

\subsubsection{Acknowledgements} This research was supported by the Ashoka Mphasis Lab - a collaboration between Ashoka University and Mphasis Limited.

%
%
%
\bibliographystyle{fkg_in_bib}
\bibliography{fkg-ref}
%





\end{document}